\documentclass[letterpaper, 10 pt, conference]{ieeeconf}  

\IEEEoverridecommandlockouts                              

\usepackage{graphicx}
\usepackage{subfig}
\usepackage{amsfonts}
\usepackage{amsmath}
\usepackage{amssymb}
\usepackage{tabularx}
\usepackage{booktabs}
\usepackage{siunitx}
\usepackage{multirow}
\usepackage{array}
\usepackage{tikz}
\usepackage{xcolor}
\usepackage{hyperref}
\usepackage{physics}
\usetikzlibrary{positioning,shapes}

\title{\LARGE \bf
LiOn-XA: Unsupervised Domain Adaptation via LiDAR-Only Cross-Modal Adversarial Training
}

\author{Thomas Kreutz$^{1}$, Jens Lemke$^{1}$, Max Mühlhäuser$^{1}$ and Alejandro Sanchez Guinea$^{1}$%
\thanks{$^{1}$The authors are with the Telecooperation Lab at the Technical University Darmstadt, Germany, {\tt\small \{<kreutz, lemke, sanchez>@tk, <max>@informatik\}.tu-darmstadt.de}}%
}

\newcolumntype{M}[1]{>{$}m{#1}<{$}}
\newcolumntype{N}{>{$}c<{$}}
\newcolumntype{C}[1]{>{$}>{\centering\let\newline\\\arraybackslash\hspace{0pt}}m{#1}<{$}}
\definecolor{vehicle}{RGB}{255, 158, 0}
\definecolor{driveable}{RGB}{0, 207, 191}
\definecolor{sidewalk}{RGB}{75, 0, 75}
\definecolor{terrain}{RGB}{112, 180, 60}
\definecolor{manmade}{RGB}{222, 184, 135}
\definecolor{vegetation}{RGB}{0, 175, 0}
\definecolor{ignore}{RGB}{0, 0, 0}

\begin{document}
\maketitle
\thispagestyle{empty}
\pagestyle{empty}

\begin{abstract}

In this paper, we propose LiOn-XA, an unsupervised domain adaptation (UDA) approach that combines \textbf{\underline{Li}}DAR-\textbf{\underline{On}}ly Cross-Modal (\textbf{\underline{X}}) learning with \textbf{\underline{A}}dversarial training for 3D LiDAR point cloud semantic segmentation to bridge the domain gap arising from environmental and sensor setup changes. Unlike existing works that exploit multiple data modalities like point clouds and RGB image data, we address UDA in scenarios where RGB images might not be available and show that two distinct LiDAR data representations can learn from each other for UDA. More specifically, we leverage 3D voxelized point clouds to preserve important geometric structure in combination with 2D projection-based range images that provide information such as object orientations or surfaces. To further align the feature space between both domains, we apply adversarial training using both features and predictions of both 2D and 3D neural networks. 
Our experiments on 3 real-to-real adaptation scenarios demonstrate the effectiveness of our approach, achieving new state-of-the-art performance when compared to previous uni- and multi-model UDA methods. Our source code is publicly available at \url{https://github.com/JensLe97/lion-xa}.

\end{abstract}


\section{Introduction}
Supervised deep learning models for LiDAR semantic segmentation trained on one dataset usually encounter domain shifts when being tested on a different dataset, which often arises from changes to geographic regions or the type of LiDAR sensor (e.g., \cite{Yi.2021}). In settings where no labels from the target domain are available, various unsupervised domain adaptation (UDA) methods have been proposed to enhance model robustness and prevent performance degradation caused by these shifts. Common UDA methods may target (i) domain-invariant data representations for deep neural networks~\mbox{(e.g., \cite{Yi.2021, Langer.2020, shaban2023lidar})} or (ii) domain-invariant feature representations using cross-modal learning strategies~\mbox{(e.g., \cite{Jaritz.2020, Jaritz.2022})}, which can be combined with adversarial training~\mbox{(e.g., \cite{Liu.2021, Peng.2021})} or contrastive learning~\mbox{(e.g., \cite{xing2023cross})}. Recent advances in learning domain-invariant feature representations from a combination of RGB and LiDAR  data have been shown to be effective for UDA. In spite of this, and especially for robotic applications, there are relevant scenarios in which RGB data might not be available or in which its usage might pose strong privacy concerns~\mbox{(e.g., \cite{muhlhauser2020street, Kreutz_2023_WACV})}.



For scenarios where only LiDAR data is available and no target domain labels can be obtained, we propose 
LiOn-XA, a UDA approach that combines
\textbf{\underline{Li}}DAR-\textbf{\underline{On}}ly Cross-Modal (\textbf{\underline{X}}) learning from two different LiDAR data representation with \textbf{\underline{A}}dversarial training. 
We motivate LiOn-XA from recent advances in LiDAR semantic segmentation, where fusing multiple LiDAR data representations helps to mitigate inherent issues of these representations~\mbox{(e.g., \cite{Cheng.2021, Tang.2020, Xu.2021b})}. 
For instance, voxelization introduces quantization loss and computation can grow cubicly with increasing resolution (limiting the receptive field), while range image-based representations distort physical dimensions (e.g., \cite{Xu.2021b}). However, voxelization preserves important geometric structure that benefits segmenting \textit{thing} classes, while range images can be processed with a much larger receptive field. 

LiOn-XA combines two LiDAR data representations with a cross-modal mimicking task (e.g., \cite{Jaritz.2020}) to obtain predictions that are more robust to a domain shift. We leverage~a)~voxelized point clouds to preserve the geometric structure, and~b)~corresponding range images for information about, for instance, the orientation and surface of objects. In addition, we align the feature spaces of the source and target domain with adversarial training. Unlike prior works that either leverage 2D features or 3D predictions~\mbox{(e.g., \cite{Liu.2021, Peng.2021})}, LiOn-XA leverages both to learn better domain-invariant feature representations. That is, during training, LiOn-XA aligns both 2D features as well as 2D and 3D predictions across domains. 

We evaluate the effectiveness of our approach on 3 real-to-real adaptation scenarios from different urban environments and sensor characteristics. Our experiments show that LiOn-XA outperforms state-of-the-art uni- and multi-modal strategies. 
Our main contributions are as follows:


\begin{itemize}
\item 
A new method for UDA that is based on cross-modal learning between two representations of the same LiDAR data, i.e., between voxelized LiDAR point clouds and their range image projections.

\item We propose a novel UDA approach called LiOn-XA, which combines LiDAR-only cross-modal learning with adversarial training to bridge country-to-country as well as sensor-to-sensor domain gaps. 
\end{itemize}

\section{Related Work}



\subsection{Domain Mapping}

Domain mapping approaches seek to obtain a common input data representation in terms of visual appearance of both the source and target domain. For instance, the works in \cite{Langer.2020, Besic.2022} address cross-sensor UDA by transforming source domain LiDAR data with a high beam size to the low resolution of the target domain. A different approach, unpaired mask transfer (UMT)~\cite{Rochan.2022} operates on range images and maps source data to the same sparsity of the target domain. Recent work in~\cite{shaban2023lidar} proposes a self-training LiDAR UDA approach that generates reliable pseudo labels for the target domain using cross-frame ensembling to aggregate predictions from multiple frames on a common input data representation of source and target domain. 

\subsection{Domain-Invariant Data Representation}

Domain-invariant data representation methods aim to learn a common input representation among both domains, which does not necessarily yields a similar visual appearance that is based on one domain. For instance, Complete\&Label~\cite{Yi.2021} address cross-sensor UDA with a shared voxel completion network, which transforms both source and target data into a common LiDAR representation. While addressing differences in sensor characteristics, such alignment strategies are not necessarily designed to bridge domain gaps for geographic-to-geographic scenarios. 

\subsection{Domain-Invariant Feature Representation}
Learning domain-invariant feature representations is usually addressed by forcing the backbones to learn a distribution over the extracted feature representation that is similar across both source and target domain. Different works proposed to learn such a feature representation by, for instance, cross-modal learning between RGB images and LiDAR point clouds~\mbox{(e.g., \cite{Jaritz.2020, Jaritz.2022})} or a combination of both cross-modal learning with adversarial training~\mbox{(e.g., \cite{Liu.2021, Peng.2021})}. Recent work in~\cite{xing2023cross} also proposed a cross-modal contrastive learning approach, which, analogous to adversarial training, aims to directly align corresponding pixel and point features across both domains. However, the assumption that RGB images synchronized with LiDAR data in both source and target domains are always available may not always hold, which limits the practicality of such methods (e.g., \cite{kong2023conda}). 

\subsection{Multi-Modal LiDAR Semantic Segmentation}
Recent state-of-the-art approaches for semantic segmentation, such as SPVNAS~\cite{Tang.2020}, (AF)\textsuperscript{2}-S3Net~\cite{Cheng.2021}, RPVNet~\cite{Xu.2021b}, 2DPASS~\cite{yan20222dpass} show that fusion-based methods that leverage LiDAR-only representations (i.e., Point Cloud, Voxels, Range Image) or LiDAR and RGB images in a cross-modal learning setting, complement each other and help mitigating issues of individual LiDAR representations to improve the overall performance. 

Compared to recent state-of-the-art LiDAR UDA methods that depend on RGB and LiDAR data, our approach especially accounts for sensor-to-sensor and country-to-country UDA scenarios where only LiDAR point cloud data is available. Different from previous works, we propose to combine cross-modal learning (e.g., \cite{Jaritz.2020}) between two different LiDAR representations with adversarial training (e.g., \cite{Liu.2021, Peng.2021}) to learn domain-invariant feature representations for LiDAR-only UDA and only depend on LiDAR data.

\section{Approach}
\begin{figure*}[t]
\smallskip
\begin{center}
    \includegraphics[width=.95\linewidth]{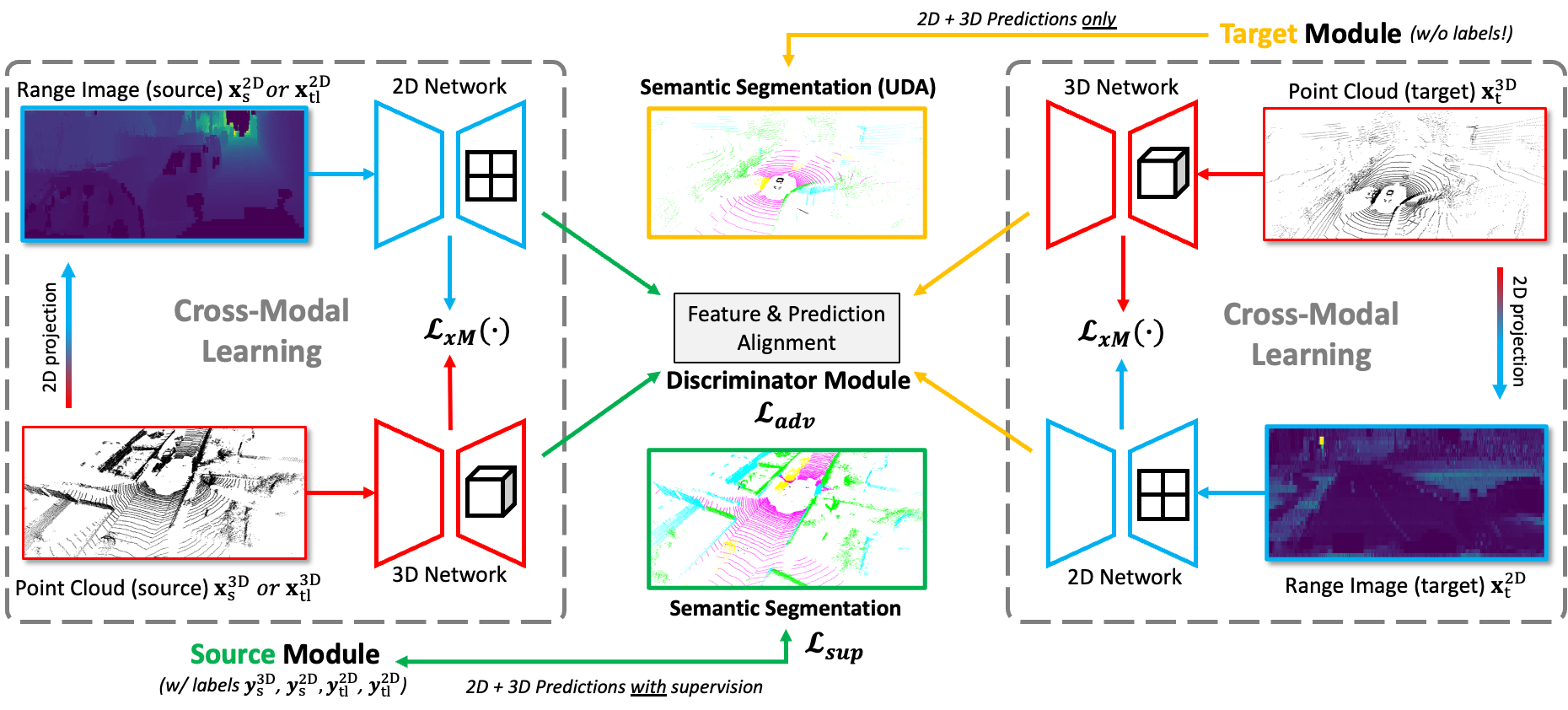}
\end{center}
    \caption[The general concept and structure of LiOn-XA]{LiOn-XA consists of a source and target module. The source module optimizes the 2D and 3D networks with a supervised segmentation loss $\mathcal{L}_{sup}$ on the source domain data as well as target-like data. The target module contains unlabelled data only from the target domain. Both modules optimize both networks with a respective cross-modal loss $\mathcal{L}_{xM}(\cdot)$. Finally, the discriminator module further connects source and target representations for unsupervised domain adaptation using an adversarial loss on the feature representations $\mathcal{L}_{adv}$ to enforce feature alignment between both domains.} 
    \label{fig:lion-xa-approachoverview}
\end{figure*}

Typically, cross-modal learning methods~(e.g., \cite{Jaritz.2020}), treat image data provided by cameras and 3D LiDAR point clouds as two distinct modalities. 
Under the assumption that 2D and 3D LiDAR data representations can learn from each other, we focus on a cross-modal learning setting for scenarios where only 3D LiDAR point cloud data is available. In this setting, we propose LiOn-XA, \textbf{\underline{Li}}DAR-\textbf{\underline{On}}ly Cross-Modal (\textbf{\underline{X}}) learning with \textbf{\underline{A}}dversarial training, where we use voxelized 3D LiDAR point clouds as a first modality and their corresponding 2D range images as a second modality.
\subsection{Overview}
\label{subsec:Architecture}

Figure~\ref{fig:lion-xa-approachoverview} depicts an overview of LiOn-XA, which consists of a source, target, and discriminator module. The source and target domain modules process the respective point clouds and perform cross-modal learning. The voxelized point clouds are processed by a 3D network, while a different 2D network processes the corresponding range-image representations. Both networks do not share features.

During training, cross-modal learning in both source and target modules enforces consistency between both network outputs, which in return improves the capabilities of both segmentation networks. However, the learning of both modalities in the source and target modules are mutually independent. We follow the idea in~\cite{Peng.2021} and use adversarial training to enforce an alignment between the features and predictions from both domains and modalities. The 2D and 3D segmentation networks serve as generators, and we include three discriminator networks inside the discriminator module for adversarial training. 
\subsection{Supervised Training}
\label{subsec:SupervisedTraining}

We train our approach on a source domain dataset $\mathcal{S}$ that consists of 3D point clouds $\boldsymbol{x}_s^{3D} \!\in\!\mathbb{R}^{N \times 3}$ with their corresponding 3D segmentation labels $\boldsymbol{y}_s^{3D} \!\in\! [\![1, C]\!]^{N}$. 

Concerning sensor-to-sensor UDA, training with additional target-like data (e.g., \cite{Langer.2020}) improves the performance of our approach (see Section~\ref{subsec:AblationStudy}) by compensating for different sensor configurations regarding beam size. In this case, we consider an additional target-like domain dataset $\mathcal{T}_{\ell}$ consisting of point clouds $\boldsymbol{x}_{tl}^{3D}$ with the associated ground truth $\boldsymbol{y}_{tl}^{3D}$. More specifically, following the approach in~\cite{Langer.2020}, the target-like point cloud data $\mathcal{T}_{\ell}$ is obtained by transforming $\boldsymbol{x}_s^{3D} \!\in\!\mathbb{R}^{N \times 3}$, so that its beam size aligns with the beam size of the target domain. We denote the corresponding range images of the source and target-like dataset as $\boldsymbol{x}_s^{2D}$ and $\boldsymbol{x}_{tl}^{2D}$.



\textit{Pixel- and Point-wise Segmentation Loss:} With provided labels, we train the 3D segmentation network in a supervised manner using the well-known cross-entropy loss:

\begin{align}
	{\mathcal{L}}_{\mathrm{seg}}^{3D}\left(\boldsymbol{x},\boldsymbol{y}\right)=-\frac{1}{N}\sum _{n=1}^{N}\sum _{c=1}^{C}w_c{\boldsymbol{y}}^{\left(n,c\right)}\log\mathbf{P}_{\boldsymbol{x}}^{\left(n,c\right)}
\end{align} 
while the 2D segmentation network is trained with:

\begin{align}
	{\mathcal{L}}_{\mathrm{seg}}^{2D}\left(\boldsymbol{x},\boldsymbol{y}\right)=-\frac{1}{H\cdot W}\sum _{h=1}^{H}\sum _{w=1}^{W}\sum _{c=1}^{C}\Bigl(w_c{\boldsymbol{y}}^{\left(h, w, c\right)} \notag \\ \log\mathbf{P}_{\boldsymbol{x}}^{\left(h, w, c\right)}\Bigl) \label{eq:Log-smoothedClassWeight}
\end{align}
where $\boldsymbol{x}$ is either 
$\boldsymbol{x}_s^{2D}$, $\boldsymbol{x}_s^{3D}$, $\boldsymbol{x}_{tl}^{2D}$, or $\boldsymbol{x}_{tl}^{3D}$, and $\boldsymbol{y}$ is equal to to the annotations $\boldsymbol{y}_s^{2D}$, $\boldsymbol{x}_{s}^{3D}$, $\boldsymbol{y}_{tl}^{2D}$ or $\boldsymbol{y}_{tl}^{3D}$. Similar to~\cite{Jaritz.2020}, we compensate for class imbalances by weighting each point by its log-smoothed classed weight~$w_c$.



Overall, the objective function for the 2D and 3D stream from the source and target-like domain reads:

\begin{small}
  
\begin{align}
\mathcal{L}_{\mathrm{sup}} = \mathcal{L}_{\mathrm{seg}}^{3D}\!\left(\mathcal{S}\right) + {\lambda}_{p}\mathcal{L}_{\mathrm{seg}}^{2D}\!\left(\mathcal{S}\right) + \mathcal{L}_{\mathrm{seg}}^{3D}\!\left(\mathcal{T}_{\ell}\right) + {\lambda}_{p}\mathcal{L}_{\mathrm{seg}}^{2D}\!\left(\mathcal{T}_{\ell}\right) 
\label{eq:SegLoss}
\end{align}
\end{small}

where $\mathcal{L}_{\mathrm{seg}}^{2D}(\cdot)$ and $\mathcal{L}_{\mathrm{seg}}^{3D}(\cdot)$ denote the loss over the whole respective datasets $\mathcal{S}$ and $\mathcal{T}_{\ell}$. We choose ${\lambda}_{p} < 1$ as a weighting hyperparameter for $\mathcal{L}_{\mathrm{seg}}^{2D}$ to limit the impact on the objective function for labels obtained by a projection in which information is lost. 



\subsection{LiDAR-Only Cross-Modal Training}
\label{subsec:Cross-ModalTraining}
The goal of cross-modal learning is to enforce consistent predictions between two modalities. Intuitively, we aim to transfer knowledge from one LiDAR data modality to the other by implementing a mimicking task. This is based on the assumption that one modality could be more sensitive to a domain shift than the other (e.g., \cite{Jaritz.2020}). For instance, point clouds lack connectivity information of an underlying surface, which is introduced by a range image projection. In this case, the more robust range image acts as a teacher to guide the more sensitive point cloud.

\begin{figure}[t]
    \centering
    \smallskip
    \smallskip
    \includegraphics[width=.95\linewidth]{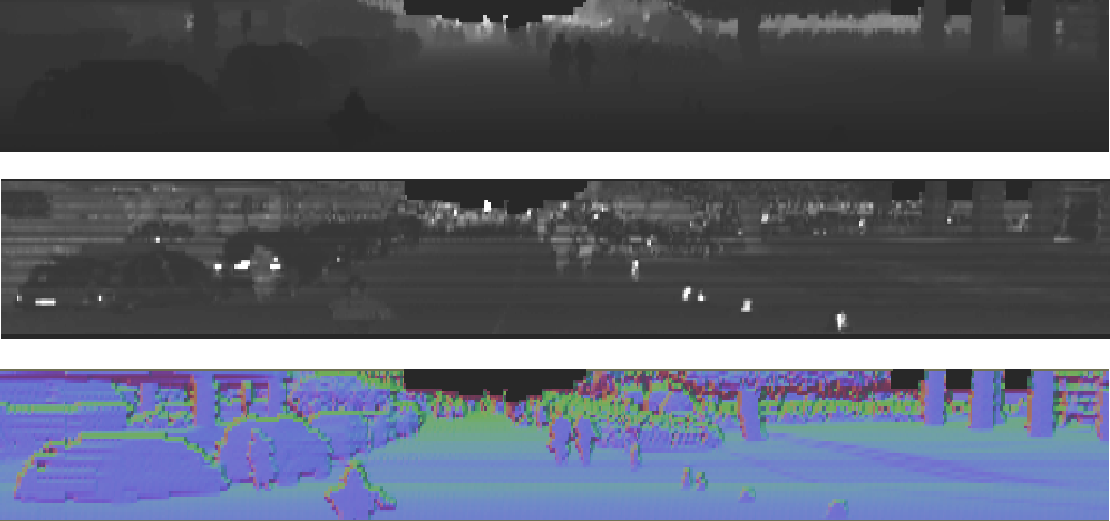}
    \caption[Range image consisting of a range, remission and normal map.]{Range image consisting of a range, remission and normal map.}
    \label{fig:RangeImageComp}
\end{figure}

\paragraph{Learning from Range Images and Point Clouds}
To complement the structural information of LiDAR point clouds, we combine them with 2D range images that we construct from the point cloud data. We hypothesize that this combination allows to extract further relevant information from the data that is more robust against domain shifts.
Figure~\ref{fig:RangeImageComp} shows a range image that has five channels, displayed as three separate images. From top to bottom, Figure~\ref{fig:RangeImageComp} consists of a range map, remission map, and normal map. By transferring the unstructured and sparse point cloud into a structured and dense range map, object contours become more visible compared to 3D point clouds. Remission values can be a major cue for the recognition of objects. For example, the high reflection of shiny \textit{thing} classes, such as cars or traffic signs, highlighted in white in the middle of Figure~\ref{fig:RangeImageComp}, facilitates the learning of these categories.

To benefit from the last component of the range image, we include normal maps as 3 separate channels for the coordinates $x$, $y$, and $z$. In contrast to other projection-based approaches (e.g., \cite{Besic.2022, Rochan.2022, Kong.2021}), we do not add the 3D coordinates directly into the images. This information is already included in the point cloud data for the 3D stream. Further, point clouds from different LiDAR sensors have differently oriented coordinate systems, which may negatively affect the adaptation process. Since the normal map values are in a fixed range and provide meaningful information about orientation, we use them as a type of canonical domain, comparable to~\cite{Yi.2021}. The model benefits from both the information related to edges and from the orientation that resembles a surface in 3D space. Point clouds lack this connectivity information of an underlying mesh~\cite{Mescheder.2019}. 

\paragraph{LiDAR-Only Cross-Modal Loss}

To learn a mapping from 2D to 3D and vice versa, we use a cross-modal loss proposed by~\cite{Jaritz.2020}, which can be formulated as:

\begin{small}
\begin{align}
	{\mathcal{L}}_{\mathrm{xM}}\left(\boldsymbol{x}\right)
	&=D_{\mathrm{KL}}\!\left(\mathbf{P}_{\boldsymbol{x}}^{\left(n,C\right)} \middle\| \mathbf{Q}_{\boldsymbol{x}}^{\left(n,C\right)}\right) \label{eq:CrossModal1}\\
	&=-\frac{1}{N}\sum _{n=1}^{N}\sum _{c=1}^{C}{\mathbf{P}}_{\boldsymbol{x}}^{\left(n,c\right)}\mathrm{log}\frac{\mathbf{P}_{\boldsymbol{x}}^{\left(n,c\right)}}{\mathbf{Q}_{\boldsymbol{x}}^{\left(n,c\right)}} \label{eq:CrossModal2}
\end{align}
\end{small}

where $D_{\mathrm{KL}}\!\left(\cdot\right)$ denotes the Kullback-Leibler divergence. In Equation~\ref{eq:CrossModal1} and Equation~\ref{eq:CrossModal2}, ${\mathbf{P}}_{\boldsymbol{x}}$ is the distribution of the main prediction that either originates from the 2D or 3D stream and ${\mathbf{Q}}_{\boldsymbol{x}}$ is the mimicry prediction that should be learned. We strive for consistency between both mimicry predictions $\mathbf{P}^{2D\rightarrow 3D}$ and $\mathbf{P}^{3D\rightarrow 2D}$.

Since the cross-modal loss is unsupervised, we apply this function to all available datasets, namely the source, target-like, and target domain. In total, the objective function from Equation~\ref{eq:SegLoss} extends to:

\begin{small}
\begin{align}
\mathcal{L}_{\mathrm{total}} = \mathcal{L}_{\mathrm{sup}} + \lambda_{s}{\mathcal{L}}_{\mathrm{xM}}\!\left(\mathcal{S}\right) + \lambda_{tl} {\mathcal{L}}_{\mathrm{xM}}\!\left(\mathcal{T}_{\ell}\right) + \lambda_{t} {\mathcal{L}}_{\mathrm{xM}}\!\left(\mathcal{T}\right) 
\label{eq:TotalLoss}
\end{align}
\end{small}

where ${\mathcal{L}}_{\mathrm{xM}}(\cdot)$ denotes the loss over the whole respective datasets $\mathcal{S}$, $\mathcal{T}_{\ell}$ or $\mathcal{T}$, with $\lambda_{s}$, $\lambda_{tl}$ and $\lambda_{t}$ being hyperparameters that weight the influence of the cross-modal loss ${\mathcal{L}}_{\mathrm{xM}}$. The target domain dataset is denoted as $\mathcal{T}$ and only contains point clouds~$\boldsymbol{x}_t^{3D}$ without annotations. 



\subsection{Discriminator Module and Adversarial Training}
\label{subsec:AdversarialTraining}

We train 3 discriminator networks to realize the adversarial domain adaptation procedure. The first discriminator~$D_{{s}^{2D}\Leftrightarrow t^{2D}}$ distinguishes between 2D source and 2D target features and outputs the probability that the input belongs to the target domain. In this setting, the 2D network as the generator is trained and updated to fool the discriminator. Likewise, the other two discriminator networks take predictions from mixed modalities and domains as input and predict the corresponding domain, i.e.,~$D_{{s}^{2D}\Leftrightarrow t^{3D}}$ and~$D_{{s}^{3D}\Leftrightarrow t^{2D}}$.

\paragraph{Discriminator Loss} We label the source domain as $0$ and the target domain as $1$, and the discriminator is trained to predict the corresponding source or target label. 
We let ${\mathcal{L}}_{\mathrm{BCE}}$ be the binary cross-entropy loss for prediction $x$ and label $y$. For the feature maps $\bar{\mathbf{F}}_{s}^{2D}$ and $\bar{\mathbf{F}}_{t}^{2D}$ from the 2D network, the discriminator loss is:
\begin{align}
	\underset{\theta_D}{\mathrm{min}}\Biggl[\frac{1}{\abs{\mathcal{S}}}\sum _{\boldsymbol{x}_{s}\in \mathcal{S}}\lambda_{D^{2D}_{sf}}{\mathcal{L}}_{\mathrm{BCE}}\!\left(D\!\left(\bar{\mathbf{F}}_{s}^{2D}\right),0\right)\notag\\+
	\frac{1}{\abs{\mathcal{T}}}\sum _{{\boldsymbol{x}}_{t}\in \mathcal{T}}\lambda_{D^{2D}_{tf}}{\mathcal{L}}_{\mathrm{BCE}}\!\left(D\!\left(\bar{\mathbf{F}}_{t}^{2D}\right),1\right)\Biggl] \label{eq:DiscrLoss}
\end{align}
where $\lambda_{D^{2D}_{sf}}$ and $\lambda_{D^{2D}_{tf}}$ are weighting hyperparameters for the source and target features $sf$ and $tf$, and $\theta_D$ are the parameters for discriminator $D$. 

\paragraph{Adversarial Objective} We update the 2D network $\theta_{2D}$ by training with the adversarial objective:

\begin{align}
	\underset{\theta_{2D}}{\mathrm{min}}\frac{1}{\abs{\mathcal{T}}}\sum _{{x}_{t}\in \mathcal{T}}\lambda_{G^{2D}_{tf}}{\mathcal{L}}_{\mathrm{BCE}}\left(D\!\left(\bar{\mathbf{F}}_{t}^{2D}\right),0\right) \label{eq:GenLoss}
\end{align}

where $\lambda_{G^{2D}_{tf}}$ is a weighting hyperparameter for the loss of generator $G$.
If we set the label of $\mathcal{L}_{\mathrm{BCE}}$ to $0$, the resulting term is~$-\log\!\left(1-D\!\left(\bar{\mathbf{F}}_{t}^{2D}\right)\right)$. The generator tries to minimize this loss, which forces the discriminator to output the wrong label for a given target domain input.

The same principle applies to the other two discriminator networks that take source and target predictions as input. The only difference is that both discriminators have the goal to align the distribution of the respective source and target domain predictions $\mathbf{P}_{t}^{2D} \Leftrightarrow \mathbf{P}_{s}^{3D}$ and $\mathbf{P}_{s}^{2D} \Leftrightarrow \mathbf{P}_{t}^{3D}$.

\subsection{Training Details}
In total, we optimize a three-fold objective loss function in the same iteration over one batch. First, we train the segmentation networks with the supervised and cross-modal loss from Equation~\ref{eq:TotalLoss}. Second, we update the segmentation networks according to Equation~\ref{eq:GenLoss}. Finally, we update the discriminators on both the source and target data using Equation~\ref{eq:DiscrLoss}. We weigh all three generator and six discriminator loss terms using hyperparameters that have been determined empirically based on prior works, which are summarized in the supplementary material. 

\section{Experiments}

We first outline our experimental setup and design. Afterward, we evaluate our approach quantitatively and qualitatively against state-of-the-art approaches on three different real-to-real domain adaptation scenarios. Furthermore, we perform an ablation study that highlights the influence of different components of our approach.

\begin{table*}[tp]
\smallskip
\smallskip
	\begin{center}
        \small
        \resizebox{0.99\textwidth}{!}{%
		\begin{tabular}{p{4.6cm}*{3}{C{1.3cm}}C{0cm}*{3}{C{1.3cm}}C{0cm}*{3}{C{1.3cm}}C{0cm}}
			\toprule
			\multicolumn{1}{c}{\multirow{2}{*}{Method}}&
			\multicolumn{3}{c}{nuScenes: USA $\rightarrow$ SG} &
			&
			\multicolumn{3}{c}{nS-Lidarseg: USA $\rightarrow$ SG} &  & \multicolumn{3}{c}{SemanticKITTI $\rightarrow$ nS-Lidarseg} & \\ \cmidrule{2-4} \cmidrule{6-8} \cmidrule{10-12}
			& \text{2D}   & \text{3D}   & \text{2D+3D} &  & \text{2D}   & \text{3D}   & \text{2D+3D} & & \text{2D}   & \text{3D}   & \text{2D+3D} & \\ \midrule
			Baseline (source only) & 53.4 & 46.5 & 61.3  &  & 58.4 & 62.8 & 68.2  &  & 47.6 & 54.9 & 61.5  &  \\ \midrule
			xMUDA~\cite{Jaritz.2020, Jaritz.2022} & 59.3 & 52.0 & 62.7  &  & 64.4 & 63.2 & 69.4  &  & \boldsymbol{57.6} & 57.7 & 63.2  &    \\
			AUDA~\cite{Liu.2021}   & 59.8 & 52.0 & 63.1  &  & -    & -    & -     &  & - & - & - & \\
			DsCML + CMAL~\cite{Peng.2021} & \boldsymbol{63.4} & 55.6 & 64.8  &  & 65.6 & 56.2 & 66.1  &  & - & - & - & \\ 
            CM-CL+PL~\cite{xing2023cross} & 63.3 & 57.1 & 66.7  &  & - & - & -  &  & - & - & - & \\ \midrule
			LiOn-XA (ours)                 & 58.0 & \boldsymbol{63.4} & \boldsymbol{68.9}  &  & \boldsymbol{67.2} & \boldsymbol{69.7} & \boldsymbol{72.8}  &  & 51.9 & \boldsymbol{70.7} & \boldsymbol{71.3}  &  \\ \midrule \midrule
			Oracle (target only)   & 66.4 & 63.8 & 71.6  &  & 75.4 & 76.0 & 79.6  &  & 75.4 & 76.0 & 79.6  & \\
			Unsupervised advantage & 4.6  & 17.3 & 7.6   &  & 8.8  & 6.9  & 4.6   &  & 4.3  & 15.8 & 9.8   & \\
			Domain gap             & 13.0 & 17.5 & 10.3  &  & 17.0 & 13.2 & 11.4  &  & 27.8 & 21.1 & 18.1  & \\
			Closed gap             & 35.4 & 97.7 & 73.8  &  & 51.8 & 52.3 & 40.4  &  & 15.5 & 74.9 & 54.1  & \\ \bottomrule
		\end{tabular}}
		\caption[Quantitative results on the nuScenes and nuScenes-Lidarseg dataset]{Quantitative results on the nuScenes, nuScenes-Lidarseg (nS-Lidarseg) and SemanticKITTI dataset from the USA to Singapore (SG). We report the mIoU for each modality (2D and 3D) as well as their ensemble (2D + 3D).}
		\label{tab:ResultsnuScenes}
	\end{center}
\end{table*}

\subsection{Datasets}
We evaluate our approach on 4 publicly available large-scale datasets that are designed for autonomous driving. More specifically, we follow related work (e.g., \cite{Jaritz.2020, Liu.2021, Peng.2021, Jiang.2021}) and use nuScenes~\cite{Caesar.2020}, nuScenes-Lidarseg~\cite{Fong.2022}, SemanticKITTI~\cite{Behley.2019}, and SemanticPOSS~\cite{Pan.2020}  
to construct country-to-country, dataset-to-dataset, and sensor-to-sensor domain adaptation scenarios. For the first scenario, we split the nuScenes dataset into LiDAR scans recorded in USA and Singapore. Since all three datasets are captured by different LiDAR sensors, we investigate two cross-sensor adaptations from SemanticKITTI to nuScenes-Lidarseg as well as from SemanticKITTI to SemanticPOSS.

\begin{table*}[t]
\smallskip
\smallskip
	\begin{center}
        \small
        \resizebox{0.99\textwidth}{!}{%
		\begin{tabular}{l*{12}{c@{\hspace{1.5\tabcolsep}}}|c}
			\toprule
			\multicolumn{1}{c}{\centering Method}                 & Person & Rider & Car & Trunk & Vegetation & Traffic-sign & Pole & Object & Building & Fence & Bike  & Ground & mIoU ($\uparrow$)  \\ 
            \midrule
			Baseline (source only) 			   & 22.77   & 1.78   & 35.91 & 16.86  & 39.84       & 7.08          & 9.73  & 0.18    & 57.03     & 1.64   & 18.17  & 41.99   & 21.08 \\ 
            \midrule
			LiDARNet~\cite{Jiang.2021}& 31.39   & $\boldsymbol{23.98}$  & 70.78 & 21.43  & 60.68       & $\boldsymbol{9.59}$          & 17.48 & $\boldsymbol{4.97}$    & 79.53     & 12.57  & $\boldsymbol{0.78}$   & 82.41   & 34.63 \\ 
            \midrule
            LiDAR-UDA~\cite{shaban2023lidar} & $\boldsymbol{65.59}$ &  2.19  & 64.12  & 27.49 & $\boldsymbol{65.40}$ & 6.44 &  $\boldsymbol{36.57}$ &  4.19 &  75.21 & $\boldsymbol{40.31}$ & 0.00 & 75.06 & $\boldsymbol{38.55}$ \\ \midrule
			LiOn-XA   (ours) & 31.86   & 13.64  & $\boldsymbol{81.30}$ & $\boldsymbol{34.49}$  & 61.86       & 6.20          & 32.25 & 0.88    & $\boldsymbol{86.35}$     & 25.44  & 0.00 & $\boldsymbol{87.43}$   & 38.48 \\
        \bottomrule
		\end{tabular}}
		\caption[Quantitative results on the SemanticKITTI to SemanticPOSS scenario]{Quantitative results on the SemanticKITTI to SemanticPOSS scenario, measured by class-wise IoU and mIoU.} 
		\label{tab:ResultsKITTI2POSS}
	\end{center}
\end{table*}

    \begin{figure*}[t]
    \captionsetup[subfloat]{farskip=0pt,captionskip=2pt}\hspace{-0.33cm}
    \subfloat[][Ground Truth.\label{fig:GroundTruth1}]{
    \begin{tikzpicture}
        \node(a){\includegraphics[width=0.17\linewidth]{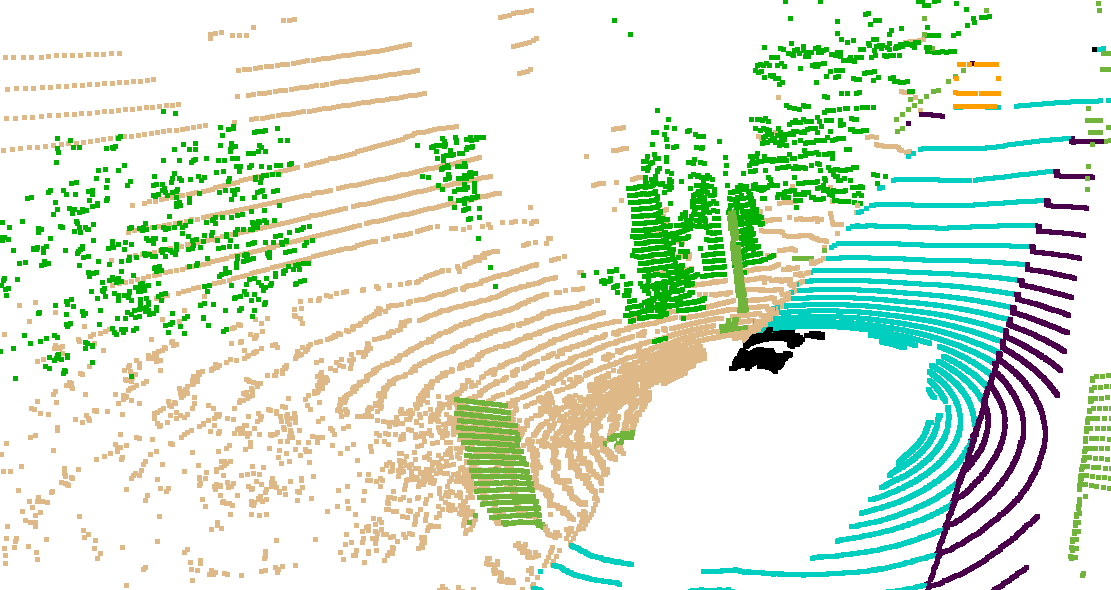}};
        \node at(a.center)[draw, red, line width=1pt, circle, minimum width=5pt, minimum height=5pt,yshift=16pt, xshift=32pt]{};
        \node at(a.center)[draw, red, line width=1pt, ellipse, minimum width=38pt, minimum height=8pt,yshift=-4pt, 	xshift=37pt, rotate=76]{};
        \node at(a.center)[draw, red, line width=1pt, ellipse, minimum width=40pt, minimum height=15pt,yshift=6pt, 	xshift=-23pt, rotate=20]{};
    \end{tikzpicture}}\hspace{0.2cm}
    \subfloat[][Baseline.\label{fig:Baseline}]{
    \begin{tikzpicture}
        \node(a){\includegraphics[width=0.17\linewidth]{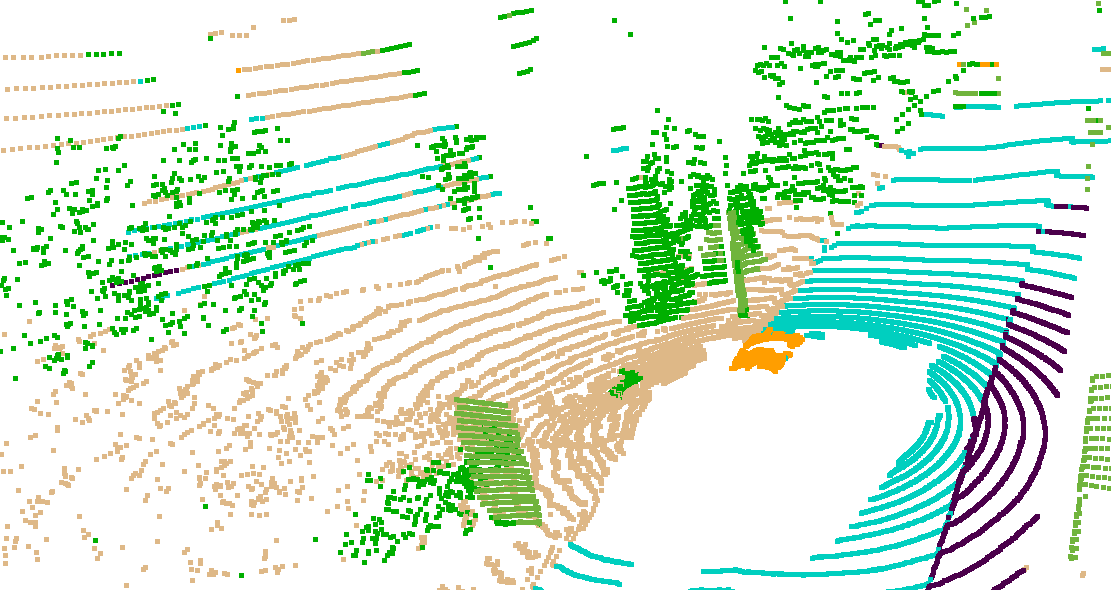}};
        \node at(a.center)[draw, red, line width=1pt, circle, minimum width=5pt, minimum height=5pt,yshift=16pt, xshift=32pt]{};
        \node at(a.center)[draw, red, line width=1pt, ellipse, minimum width=38pt, minimum height=8pt,yshift=-4pt, 	xshift=37pt, rotate=76]{};
        \node at(a.center)[draw, red, line width=1pt, ellipse, minimum width=40pt, minimum height=15pt,yshift=6pt, 	xshift=-23pt, rotate=20]{};
    \end{tikzpicture}}\hspace{0.2cm}
    \subfloat[][LiOn-XA (2D).\label{fig:ADALS_2D}]{
    \begin{tikzpicture}
        \node(a){\includegraphics[width=0.17\linewidth]{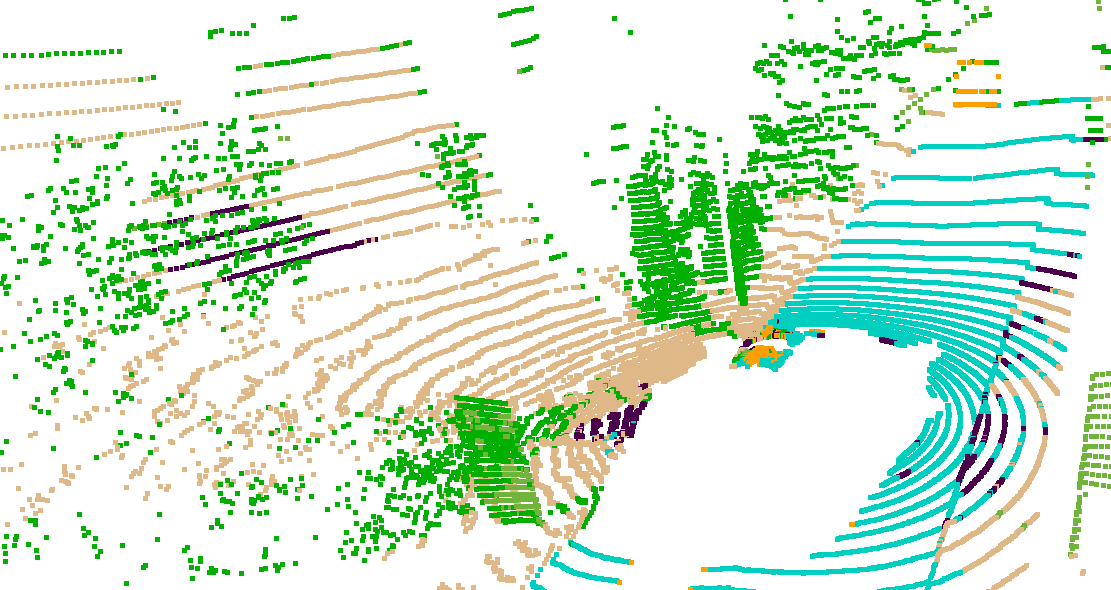}};
        \node at(a.center)[draw, red, line width=1pt, circle, minimum width=5pt, minimum height=5pt,yshift=16pt, xshift=32pt]{};
        \node at(a.center)[draw, red, line width=1pt, ellipse, minimum width=38pt, minimum height=8pt,yshift=-4pt, 	xshift=37pt, rotate=76]{};
        \node at(a.center)[draw, red, line width=1pt, ellipse, minimum width=40pt, minimum height=15pt,yshift=6pt, 	xshift=-23pt, rotate=20]{};
    \end{tikzpicture}}\hspace{0.2cm}
    \subfloat[][LiOn-XA (3D).\label{fig:ADALS_3D}]{
    \begin{tikzpicture}
        \node(a){\includegraphics[width=0.17\linewidth]{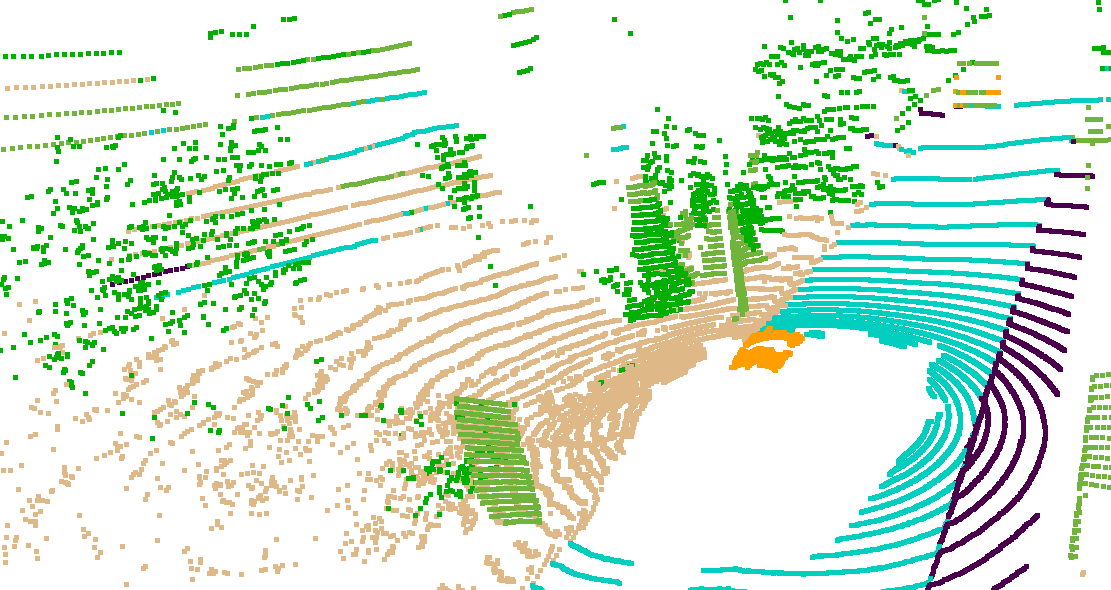}};
        \node at(a.center)[draw, red, line width=1pt, circle, minimum width=5pt, minimum height=5pt,yshift=16pt, xshift=32pt]{};
        \node at(a.center)[draw, red, line width=1pt, ellipse, minimum width=38pt, minimum height=8pt,yshift=-4pt, 	xshift=37pt, rotate=76]{};
        \node at(a.center)[draw, red, line width=1pt, ellipse, minimum width=40pt, minimum height=15pt,yshift=6pt, 	xshift=-23pt, rotate=20]{};
    \end{tikzpicture}}\hspace{0.2cm}
    \subfloat[][LiOn-XA (2D + 3D).\label{fig:ADALS}]{
    \begin{tikzpicture}
        \node(a){\includegraphics[width=0.17\linewidth]{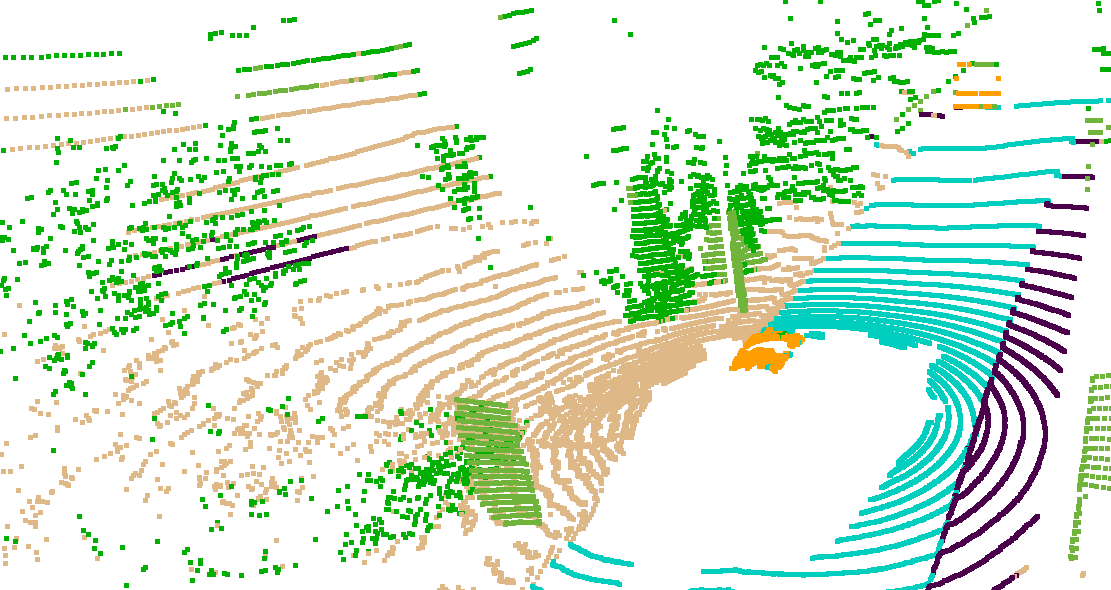}};
        \node at(a.center)[draw, red, line width=1pt, circle, minimum width=5pt, minimum height=5pt,yshift=16pt, xshift=32pt]{};
        \node at(a.center)[draw, red, line width=1pt, ellipse, minimum width=38pt, minimum height=8pt,yshift=-4pt, 	xshift=37pt, rotate=76]{};
        \node at(a.center)[draw, red, line width=1pt, ellipse, minimum width=40pt, minimum height=15pt,yshift=6pt, 	xshift=-23pt, rotate=20]{};
    \end{tikzpicture}}
    \caption[Qualitative results on the SemanticKITTI to nuScenes-Lidarseg adaptation scenario]{Qualitative results on the SemanticKITTI to nuScenes-Lidarseg adaptation scenario. \textcolor{vehicle}{$\blacksquare$} Vehicle, \textcolor{driveable}{$\blacksquare$}~Driveable surface, \textcolor{sidewalk}{$\blacksquare$} Sidewalk, \textcolor{terrain}{$\blacksquare$} Terrain, \textcolor{manmade}{$\blacksquare$} Manmade, \textcolor{vegetation}{$\blacksquare$}~Vegetation, \textcolor{ignore}{$\blacksquare$} Ignore label.
     }\label{fig:QualitativeResults}
\end{figure*}

\subsection{Implementation Details}
We use the official SalsaNext~\cite{Cortinhal.2020} implementation as 2D semantic segmentation network with 32~feature channels. Similar to~\cite{Jiang.2021}, we replace all batch normalization layers with instance normalization for better conversion and performance results of the generator module. 
We cut the \ang{360} range image and randomly take a $512$ pixel wide cutout with the same height as the target beam size. Therefore, we are able to utilize data from all viewpoints and are not limited to a front camera view as in~\cite{Jaritz.2020}. Further, we use random horizontal flips and randomly remove parts of the range image as a data augmentation. 
Finally, we normalize each channel of the range image by subtracting the mean and dividing by the standard deviation of the source dataset. 

To further facilitate the adaptation process, we align the height $H$ and width $W$ of the source range images to the dimensions of the target domain. Other works have aligned the target range images to the source domain or to the one with a smaller height~\cite{Jiang.2021}. We argue that a transformation to the target domain benefits the learning process, as no ground truth is available in this domain.

For our 3D stream segmentation network, we use the official implementation of SparseConvNet~\cite{Graham.2018} with a U-Net architecture. In the same way as~\cite{Jaritz.2020}, we implement the submanifold sparse convolution with 16~U-Net features and 6 times downsampling. To obtain a voxelized grid with only one 3D point per voxel, we set the voxel size to \SI{5}{\cm}. For the 3D data augmentation, we implement random rotation, translation, and flipping of the $x$- and $y$-axis.

\paragraph{Training}
Our models are trained from scratch on a single NVIDIA Tesla V100S (PCIe) with $32$~GB RAM 
with a batch size of $8$ for a maximum number of iterations of 100k.
In each iteration, we accumulate the gradients of the source, target-like, and target batch to jointly train both modality networks. The values for the hyperparameters in our loss functions have been selected empirically and vary across the different UDA evaluation scenarios. For the sake of reproducibility, we provide a full list of all hyperparameters in the supplementary material. 
We optimize the 2D segmentation network with stochastic gradient descent (SGD), an initial learning rate of $2.5 \times$ $10^{-3}$, and momentum $0.9$.

The 3D segmentation network is trained using ADAM with a learning rate of $10^{-3}$, $\beta_1 = 0.9$, and $\beta_2 = 0.999$. For both the 2D and the 3D training procedure, we choose a Multi Step Learning rate scheduler with $\gamma = 0.1$ and milestones at iterations 80k and 90k at which the learning rate is multiplied by $\gamma$. 

The discriminator networks are optimized with ADAM, a learning rate of $10^{-4}$, $\beta_1 = 0.9$, and $\beta_2 = 0.99$. We schedule the learning rate by following a polynomial annealing procedure as used in~\cite{Chen.2018}. In this policy learning rate is multiplied by $\left(1 - \frac{iter}{max\_iter}\right)^{power}$ where $iter$ is the current iteration, $max\_iter$ is the maximum number of iterations, and $power$ is the ``poly'' power that we set to $0.9$. 

\paragraph{Metrics}
We evaluate the performance of our approach with the well-known intersection-over-union (IoU) and corresponding mean IoU (mIoU) metric.
The mIoU is reported separately for both the 2D and 3D segmentation models. 
The model for each modality is chosen independently based on their checkpoints achieving the best mIoU on the validation set.  
Finally, we report the softmax average over both probability distributions as in~\cite{Jaritz.2020}.
\subsection{Quantitative Results}
The mIoU results for the USA to Singapore (SG) and the SemanticKITTI to nuScenes-Lidarseg domain adaptation scenarios are summarized in Table~\ref{tab:ResultsnuScenes}. 
We evaluate LiOn-XA against four state-of-the-art methods xMUDA~\cite{Jaritz.2020}, AUDA~\cite{Liu.2021}, DsCML + CMAL~\cite{Peng.2021}, and CM-CL+PL~\cite{xing2023cross}. The baseline is trained only on the source domain, while the oracle is trained only on the target domain. At the bottom, we further report the difference between LiOn-XA and the baseline (unsupervised advantage), the difference between oracle and baseline performance (domain gap), and how much of the domain gap has been closed by applying LiOn-XA (closed gap, i.e., $\dfrac{IoU(\text{LiOn-XA}) - IoU(\text{baseline})}{IoU(\text{oracle}) - IoU(\text{baseline})} \cdot 100$).

From Table~\ref{tab:ResultsnuScenes}, we observe that LiOn-XA outperforms the baseline by a large margin. The ensemble (2D + 3D) achieves the best results with an increase of $2.2$~mIoU compared to CM-CL+PL. 
In addition, LiOn-XA outperforms xMUDA on the 3D and the combined stream of the SemanticKITTI to nuScenes-Lidarseg scenario. As expected, the combination of both modalities achieves the highest scores, which compensates the lower 2D performance. The results in Table~\ref{tab:ResultsnuScenes} verify the primary assumption of our approach: 2D and 3D LiDAR representations are complementary in cross-modal learning setting and can learn robust domain-invariant features for UDA when combined with adversarial training.

For the SemanticKITTI to SemanticPOSS scenario, we compare LiOn-XA to the uni-modal approach LiDARNet~\cite{Jiang.2021} and the recently proposed self-training approach LiDAR-UDA~\cite{shaban2023lidar}. Table~\ref{tab:ResultsKITTI2POSS} summarizes the IoU results for each of the 12 classes in addition to the mIoU. LiOn-XA reaches an overall performance of~$38.48$~mIoU, which is an improvement over LiDARNet and a performance comparable to LiDAR-UDA with $38.55$. The results show that the adaptation capability of LiOn-XA is superior in many classes. We hypothesize that the additional geometric information in LiOn-XA from the 3D stream helps to learn a high-level, domain-invariant feature representation that transfers well between both datasets for classes, such as ``Car'', ``Building'', and ``Ground''. 
However, the decrease in performance in both our approach and LiDAR-UDA compared to the baseline for the ``Traffic-sign'' and the ``Bike'' classes likely indicates a considerable change in the data distribution of these classes. The difference is more severe for the class ``Bike'' as LiOn-XA seems not to be able to adapt to the large number of bikes in the campus scenes of SemanticPOSS. 

\subsection{Ablation Study}
\label{subsec:AblationStudy}
We conduct ablation studies regarding adding the discriminator module and target-like data to LiOn-XA. The results are summarized in Table~\ref{tab:AblationStudy}. Without the discriminator module (Dis), the overall performance drops from $68.9$ to $66.4$ mIoU for the nuScenes dataset. In particular, the adversarial training technique achieves consistent improvements concerning the 3D network. The domain gap in this setting is mostly due to environmental changes and the 2D network falls slightly behind when aligning the source and target features. We argue that the combination of 2D source and 3D target or vice versa exchanges feature information from both domains, in which case the 3D network benefits.

For the SemanticKITTI to nuScenes-Lidarseg scenario (right-hand side Table~\ref{tab:AblationStudy}), we include target-like data (Tgl) based on~\cite{Langer.2020} into the learning procedure. It can be observed that a discriminator module alone results in a major performance gain for all three modality options due to enforcing a domain-invariant feature space. Moreover, adding an auxiliary loss function with target-like data in addition to the discriminators further improves the performance of both network streams. Our model benefits from the domain mapping procedure into a representation that approximates the visual appearance of the target dataset.

\subsection{Qualitative Results}
\begin{table}[t]
	\begin{center}
            \large
            \resizebox{\linewidth}{!}{%
		\begin{tabular}{p{4cm}*{3}{C{1.3cm}}C{0cm}*{3}{C{1.3cm}}C{0cm}}
			\toprule
			\multicolumn{1}{c}{\multirow{2}{*}{Method}}&
			\multicolumn{3}{c}{nuScenes: USA $\rightarrow$ SG} &
			&
			\multicolumn{3}{c}{SemanticKITTI $\rightarrow$ nS-Lidarseg} & \\ \cmidrule{2-4} \cmidrule{6-8}
			& \text{2D}   & \text{3D}   & \text{2D+3D} &  & \text{2D}   & \text{3D}   & \text{2D+3D} & \\ \midrule
			LiOn-XA (w/o Dis)      & \boldsymbol{58.2} & 60.9 & 66.4  &  & 40.9 & 60.9 & 61.9  &  \\
			LiOn-XA (w/ Dis)       & 58.0 & \boldsymbol{63.4} & \boldsymbol{68.9}  &  & 51.6 & 70.4 & 71.0  &  \\
			LiOn-XA(w/ Dis+Tgl) & -    & -    & -     &  & \boldsymbol{51.9} & \boldsymbol{70.7} & \boldsymbol{71.3}  &  \\ \bottomrule
		\end{tabular}}
		\caption[Influence of discriminator and target-like data stream components]{Influence of discriminator (Dis) stream and target-like (Tgl) data. We report the mIoU for each modality.}
		\label{tab:AblationStudy}
	\end{center}
\end{table}

We qualitatively evaluate our approach against the baseline in Figure~\ref{fig:QualitativeResults}. We compare the ground truth labels in Figure~\ref{fig:GroundTruth1} with the baseline approach in Figure~\ref{fig:Baseline}, which is trained only on the source domain without a domain adaptation strategy. In addition, we show LiOn-XA as the 2D + 3D predictions in Figure~\ref{fig:ADALS}. The 2D model in Figure~\ref{fig:ADALS_2D} is able to detect the vehicle in the upper part of the point cloud, whereas the 3D model in Figure~\ref{fig:ADALS_3D} is uncertain and primarily predicts vegetation. In this case, the 2D segmentation output is more robust to domain changes, which benefits the predictions of their ensemble (2D + 3D). For other parts of the point cloud (red ellipsis on the right hand side), the 3D modality is stronger at predicting the sidewalk class. It becomes evident that both representations complement each other and their combination yields the best possible results from both.  

\section{Conclusion}

We propose LiOn-XA, a new UDA approach for 3D LiDAR point cloud semantic segmentation. LiOn-XA uses LiDAR-only cross-modal learning, where two different LiDAR representations learn from each other to mitigate the negative effects of domain shifts in sensor-to-sensor and cross-city domain adaptation. 
For LiDAR-only cross-modal learning, we propose combining a cross-modal mimicking task, adversarial training, and target-like data generated from the source domain. Experiments on three unsupervised domain adaptation scenarios show that LiOn-XA can successfully improve the performance over non-adaptive baselines as well as multi-modal state-of-the-art works. In the future, our approach could be applied to different modalities other than LiDAR, such as radar data.


\section*{Acknowledgement}
\small
This work has been partially funded by the LOEWE initiative (Hesse, Germany) within the emergenCITY centre and by the Federal Ministry of Education and Research (BMBF) grant 01$|$S17050. 






\bibliographystyle{IEEEtran}
\bibliography{IEEEabrv,IEEEexample}

\appendix

\subsection{LiDAR-Only Cross-Modal Learning Evaluation}
We analyze the domain adaptation scenario from USA to Singapore on the nuScenes-Lidarseg~\cite{Fong.2022} dataset to confirm our hypothesis on the effectiveness of the LiDAR-only cross-modal learning. Table~\ref{tab:ResultsClassesNuScenes-Lidarseg} shows the IoU class scores for each modality and the softmax average. We list the number of points for each class in the target domain test set. Based on the best results of both the 2D and 3D stream, we can infer that the two representations complement each other. For example, the 2D network performs slightly better on the ``Sidewalk'' class and achieves a higher result on the  ``Terrain'' class. In combination with the 3D network, the performance is improved in almost all cases, which justifies the usage of both networks. 

We observe that the ensemble IoU (2D+3D) per class seems to correlate with the number of points in that class. The 2D+3D IoU for ``Driveable surface'' reaches the highest IoU of $93.0$, while having the greatest number of points, i.e., $24.7$ million. LiOn-XA performs similarly for the ``Vegetation'', ``Manmade'', and ``Terrain'' class that are represented by $16.7$, $13.9$, and $12.1$ million points. With only $6.1$ million points, the ``Sidewalk'' class is harder to classify. However, there is an exception to this hypothesis. Despite being represented by merely $2.6$~million points, LiOn-XA can achieve a high IoU of $84.4$. We hypothesize that structural differences between background and foreground classes provide sufficient cues for the semantic segmentation model. In general, the class distributions from the USA compared to Singapore are similar for the most part~\cite{Kong.2021}. The intra-domain distributions are imbalanced, which impacts the segmentation performance as smaller classes are harder to detect.    

\begin{table}[ht]
	\begin{center}
        \large
            \resizebox{\linewidth}{!}{%
		\begin{tabular}{l*{3}{C{1.5cm}}C{2.5cm}}
			\toprule
			\multicolumn{1}{c}{\multirow{2}{*}{\shortstack[c]{Class}}} & \multicolumn{3}{c}{nuScenes-Lidarseg: USA $\rightarrow$ SG}  & \multirow{2}{*}{\shortstack[c]{\# of points\\in millions}}  \\ \cmidrule{2-4}
			\multicolumn{1}{c}{}                       & \text{2D}   		& \text{3D}   		& \text{2D+3D}     &       \\ \midrule
			Vehicle                                    & 65.7               & \boldsymbol{81.3}     & 84.4      & 2.6                                             \\
			Driveable surface                          & 90.2               & \boldsymbol{91.5}     & 93.0      & 24.7                                            \\
			Sidewalk                                   & \boldsymbol{41.3}      & 40.3              & 45.6      & 6.1                                             \\
			Manmade                                    & 63.5               & \boldsymbol{74.7}     & 74.2      & 13.9                                            \\
			Terrain                                    & \boldsymbol{61.6}      & 52.8              & 63.2      & 12.1                                            \\
			Vegetation                                 & 68.4               & \boldsymbol{77.8}     & 76.2      & 16.7                                            \\ \bottomrule
		\end{tabular}}
		\caption[Semantic segmentation performance on the USA to Singapore (SG) adaptation scenario]{Semantic segmentation performance of LiOn-XA on the USA to Singapore adaptation scenario based on nuScenes-Lidarseg. We report the IoU of each class for the 2D, 3D, and combined modality as well as the number of points in the target domain test set. The best result of both network streams is marked in bold.}
		\label{tab:ResultsClassesNuScenes-Lidarseg}
	\end{center}
\end{table}

\subsection{More Detailed Architecture of LiOn-XA}

\begin{figure*}[t]
\begin{center}
    \includegraphics[width=\linewidth]{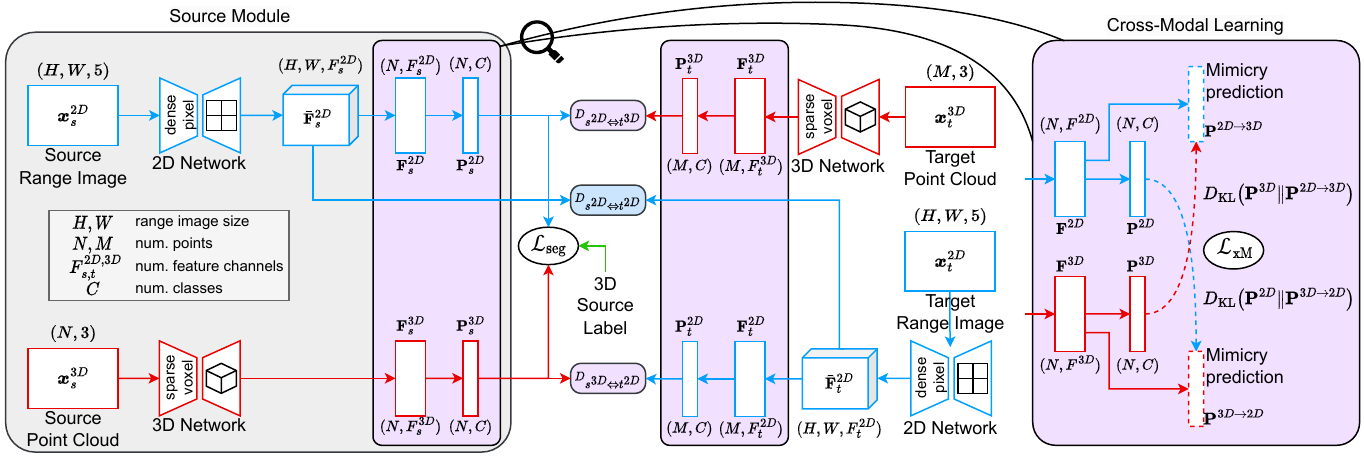}
\end{center}
    \caption[The basic architecture of LiOn-XA]{The basic architecture of LiOn-XA. It is best viewed from both sides to the middle following the data flows. } 
    \label{fig:lion-xa-architectureoverview}
\end{figure*}

A more detailed network architecture of our approach is summarized in Figure~\ref{fig:lion-xa-architectureoverview}. The flow of the 3D modality appears in red, and the 2D flow is depicted in blue.



3D point clouds from the source and target domain are denoted as $\boldsymbol{x}_{s}^{3D}\!\in\! \mathbb{R}^{N \times 3}$ and $\boldsymbol{x}_{t}^{3D}\!\in\! \mathbb{R}^{M \times 3}$, where $N$ and $M$ specify  the number of points, and usually $N \neq M$.
A shared 3D segmentation network independently processes both $\boldsymbol{x}_{s}^{3D}\!\in\! \mathbb{R}^{N \times 3}$ and $\boldsymbol{x}_{t}^{3D}\!\in\! \mathbb{R}^{M \times 3}$, which yields two different feature maps of shape $(N, F_s^{3D})$ and $(M, F_t^{3D})$, respectively. Next, both feature maps are used to classify each point, which results in softmax predictions $\mathbf{P}_s^{3D}$ and $\mathbf{P}_t^{3D}$ over all classes.

Similar to the 3D data, we define \hbox{$\boldsymbol{x}_{s}^{2D}\!\in\! \mathbb{R}^{H \times W \times 5}$} and $\boldsymbol{x}_{t}^{2D}\!\in\! \mathbb{R}^{H \times W \times 5}$ as source and target range images, which are processed by a shared 2D segmentation network independently. To account for cross-modal learning between the 2D and 3D modalities, we map the features of each pixel back to point features using a feature lifting process. More specifically, we transform the 2D feature maps $\bar{\mathbf{F}}_s^{2D}$, $\bar{\mathbf{F}}_t^{2D}$ from shape $(H, W, F_s^{2D})$ and $(H, W, F_t^{2D})$ to sparse feature maps $\mathbf{F}_{s}^{2D}$, $\mathbf{F}_{t}^{2D}$ of shape $(N, F_t^{3D})$ and $(M, F_t^{3D})$. Analogous to the 3D stream, we obtain the segmentation predictions $\mathbf{P}_s^{2D}$ and $\mathbf{P}_t^{2D}$.



To mimic predictions between both LiDAR data representations, we utilize cross-modal learning as proposed in~\cite{Jaritz.2020}. The cross-modal module is summarized on the right-hand side of Figure~\ref{fig:lion-xa-architectureoverview}, which is applied to both domains. 
Given the feature maps $\mathbf{F}^{2D}$ and $\mathbf{F}^{3D}$ from the 2D and 3D stream (independently of the domain), the last linear layer of the network outputs the main predictions $\mathbf{P}^{2D}$ and $\mathbf{P}^{3D}$. These probabilities result from a softmax function over the segmentation logits. 


To improve the UDA capability, we use adversarial training techniques in addition to cross-modal learning. Similar to~\cite{Peng.2021}, we use two discriminator networks that try to distinguish source from target predictions. Across both modalities, the 2D source and 3D target predictions are fed into discriminator $D_{s^{2D}\Leftrightarrow t^{3D}}$, whereas the 3D source and 2D target predictions are handled by discriminator $D_{s^{3D}\Leftrightarrow t^{2D}}$. It has been shown in~\cite{Peng.2021} that the combination of both modalities provides the best performance compared to single-stream predictions. 
To enforce a domain-invariant feature representation, we add a third domain discriminator $D_{s^{2D}\Leftrightarrow t^{2D}}$ in a similar fashion to~\cite{Liu.2021}, which discriminates the feature vectors of the 2D network from source and target domain.

\section{Loss Function - Weighting Hyperparameters}
The weighting hyperparameters for the cross-modal loss and the adversarial losses are summarized in Table~\ref{tab:Hyperparameters}, which were optimized empirically for each domain adaptation scenario using a small grid search among similar values used in related works~\cite{Jaritz.2020, Peng.2021, Liu.2021}. 
	
\begin{table*}[t]
    \begin{center}
    \small
    \resizebox{0.55\linewidth}{!}{%
        \begin{tabular}{c*{4}{N}}
            \toprule
            \multicolumn{1}{c}{Hyperparamter} &
            \begin{tabular}[c]{@{}c@{}}nuScenes:\\ USA $\rightarrow$ SG \end{tabular} &
            \begin{tabular}[c]{@{}c@{}}nS-Lidarseg:\\ USA $\rightarrow$ SG\end{tabular} &
            \begin{tabular}[c]{@{}c@{}}SemanticKITTI \\ $\rightarrow$ nS-Lidarseg\end{tabular} &
            \begin{tabular}[c]{@{}c@{}}SemanticKITTI \\ $\rightarrow$ POSS\end{tabular} \\ \midrule
            $\lambda_s$             & 0.8  & 0.8  & 0.1   & 0.8   \\
            $\lambda_{tl}$          & 0    & 0    & 0.02  & 1.0   \\
            $\lambda_t             $& 0.1  & 0.1  & 0.01  & 0.1   \\
            $\lambda_p             $& 0.5  & 0.8  & 0.1   & 0.8   \\
            $\lambda_{G^{2D}_{tp}} $& 0.07 & 0.07 & 0.07  & 0.07\\
            $\lambda_{G^{3D}_{tp}} $& 0.05 & 0.05 & 0.05  & 0.05  \\
            $\lambda_{G^{2D}_{tf}} $& 0.02 & 0.07 & 0.001 & 0.001 \\
            $\lambda_{D^{2D}_{tp}} $& 0.2  & 0.2  & 0.2   & 0.2   \\
            $\lambda_{D^{3D}_{tp}} $& 0.2  & 0.2  & 0.2   & 0.2   \\
            $\lambda_{D^{2D}_{tf}} $& 0.2  & 0.2  & 0.2   & 0.05  \\
            $\lambda_{D^{2D}_{sp}} $& 0.1  & 0.1  & 0.1   & 0.1   \\
            $\lambda_{D^{3D}_{sp}} $& 0.1  & 0.1  & 0.1   & 0.1   \\
            $\lambda_{D^{2D}_{sf}} $& 0.1  & 0.1  & 0.1   & 0.05  \\ \bottomrule
        \end{tabular}}
        \caption[Value assignments for the weighting hyperparameters]{Value assignments for the weighting hyperparameters of each domain adaptation scenario.}
        \label{tab:Hyperparameters}
    \end{center}
\end{table*}

\section{Dataset Splits}
\subsection{USA to Singapore (nuScenes and nuScenes-Lidarseg)}
\label{subsec:USAtoSingapore}
This adaptation scenario for the nuScenes~\cite{Caesar.2020} and nuScenes-Lidarseg~\cite{Fong.2022} dataset covers challenges caused by changes in the geographic scene layout. For instance, the urban street scenes appear differently due to the rule change of right to left-hand-side driving. Additionally, the shape of objects can vary from country to country.

For this experiment, we split the data into~15,695~source domain scans and 15,364 target domain scans following the setup in~\cite{Jaritz.2020}. The target domain data is further divided into~9,665~training, 2,770~validation, and~2,929~test frames. We use the metadata provided by the nuScenes dataset to separate the domains based on the location.

\paragraph{nuScenes Categories.} Due to the scarcity of some class objects, we merge the classes from nuScenes into five remaining categories: \textbf{vehicle} (car, truck, bus, trailer, and construction vehicle), \textbf{pedestrian}, \textbf{bike} (motorcycle and bicycle), \textbf{traffic boundary} (traffic cone and barrier), and \textbf{background}.
\paragraph{nuScenes-Lidarseg Categories.} Since the nuScenes-Lidarseg dataset subdivides the background into more accurate semantic classes, we decide on six resulting categories for this adaptation scenario. We group the classes car, truck, bus, trailer, construction vehicle, bicycle, and motorcycle under vehicle. The remaining five categories are subcategories of the background class: driveable surface, sidewalk, manmade, terrain, and vegetation.

\subsection{SemanticKITTI to nuScenes-Lidarseg}
\label{subsec:SemanticKITTItoNuscenes}
The domain shift in the scenario from SemanticKITTI~\cite{Behley.2019} to nuScenes-Lidarseg is caused by environmental changes and especially by a LiDAR sensor change. First of all, the beam size drops from 64 to 32 channels, which halves the factor of detail in the unlabeled target domain. The general appearance of the target point clouds is much sparser and provides fewer spatial cues. As a consequence, we assume that smaller objects with a limited number of points are harder to recognize, which has been shown in~\cite{Wang.2020}. 
Furthermore, the vertical field of view~(FOV) comprises different parts of the surroundings. Due to the wider FOV in nuScenes-Lidarseg, there are areas in these scans that the LiDAR sensor of SemanticKITTI cannot capture. Apart from the mounting position, the sampling pattern differs w.r.t. the vertical angular resolution. For instance, the LiDAR beams in nuScenes-Lidarseg are linearly distributed, whereas the resolution of the 64-beam Velodyne sensor decreases above and below the center. 

In this domain adaptation scenario, we use scenes 0, 1, 2, 3, 4, 5, 6, 9, and 10 as the training set for SemanticKITTI, which consist of 18,029 scans. We keep the official training set of nuScenes-Lidarseg and use the validation set as our test set. 

As we cover closed set domain adaptation, we select six shared categories by merging them and ignoring minor classes. 
The exact mapping of the SemanticKITTI and nuScenes-Lidarseg classes is summarized in Table~\ref{tab:MappingKITTI-nuScenes}. After class mapping, we end up with the same six classes as in our USA to Singapore domain adaptation scenario (cf., Section~\ref{subsec:USAtoSingapore}).

\subsection{SemanticKITTI to SemanticPOSS}
\label{subsec:SemanticKITTItoPOSS}
In the domain adaptation scenario from SemanticKITTI to SemanticPOSS~\cite{Pan.2020}, the domain shift is primarily caused by the appearance of different cities and changes in sensor characteristics. The discrepancy between the number of beams is lower than for the SemanticKITTI to nuScenes-Lidarseg scenario, i.e., 64 to 40 beams compared to 64 to 32 beams. However, the FOV does not match and the beam angles are non-linearly distributed. Overall, the domain gap in this scenario is challenging, most importantly because of the dynamic and crowded scenes in the target domain.

We split the SemanticPOSS dataset into sequences 0, 1, 2, and 3 as training set,~4 as validation set, and 5 as test set. This results in 1,988 scans for training and 500 scans for validation and testing, respectively. The training set for SemanticKITTI is the same as for the nuScenes-Lidarseg domain adaptation. 

We follow the work of~\cite{Jiang.2021} and apply the same class mappings, which results in 12 classes: person, bicyclist, car, trunk, vegetation, traffic sign, pole, object, building, fence, bike, and ground. The mapping process is summarized in Table~\ref{tab:MappingKITTI-POSS}. 

\begin{table*}[t]
    \begin{center}
    \small
        \resizebox{0.7\linewidth}{!}{%
        \begin{tabular}{ll|ll}
            \toprule
            SemanticKITTI class  & Mapped class      & nuScenes-Lidarseg class & Mapped class      \\ \midrule
            unlabeled            & ignore            & ignore                  & ignore            \\
            outlier              & ignore            & barrier                 & ignore            \\
            car                  & vehicle           & bicycle                 & vehicle           \\
            bicycle              & vehicle           & bus                     & vehicle           \\
            bus                  & ignore            & car                     & vehicle           \\
            motorcycle           & vehicle           & construction-vehicle    & vehicle           \\
            on-rails             & ignore            & motorcycle              & vehicle           \\
            truck                & vehicle           & pedestrian              & ignore            \\
            other-vehicle        & ignore            & traffic-cone            & ignore            \\
            person               & ignore            & trailer                 & vehicle           \\
            bicyclist            & vehicle           & truck                   & vehicle           \\
            motorcyclist         & vehicle           & driveable-surface       & driveable-surface \\
            road                 & driveable-surface & other-flat              & ignore            \\
            parking              & driveable-surface & sidewalk                & sidewalk          \\
            sidewalk             & sidewalk          & terrain                 & terrain           \\
            other-ground         & ignore            & manmade                 & manmade           \\
            building             & manmade           & vegetation              & vegetation        \\
            fence                & manmade           &                         &                   \\
            other-structure      & ignore            &                         &                   \\
            lane-marking         & driveable-surface &                         &                   \\
            vegetation           & vegetation        &                         &                   \\
            trunk                & vegetation        &                         &                   \\
            terrain              & terrain           &                         &                   \\
            pole                 & manmade           &                         &                   \\
            traffic-sign         & manmade           &                         &                   \\
            other-object         & manmade           &                         &                   \\
            moving-car           & vehicle           &                         &                   \\
            moving-bicyclist     & vehicle           &                         &                   \\
            moving-person        & ignore            &                         &                   \\
            moving-motorcyclist  & vehicle           &                         &                   \\
            moving-on-rails      & ignore            &                         &                   \\
            moving-bus           & ignore            &                         &                   \\
            moving-truck         & vehicle           &                         &                   \\
            moving-other-vehicle & ignore            &                         &                   \\ \bottomrule
        \end{tabular}}
        \caption[Class mapping for the SemanticKITTI and nuScenes-Lidarseg dataset]{Class mapping for the SemanticKITTI and nuScenes-Lidarseg dataset.}
        \label{tab:MappingKITTI-nuScenes}
    \end{center}
\end{table*}

\begin{table*}[t]
    \begin{center}
    \small
        \resizebox{0.7\linewidth}{!}{%
        \begin{tabular}{ll|ll}
            \toprule
            SemanticKITTI class  & Mapped class & SemanticPOSS class & Mapped class \\ \midrule
            unlabeled            & ignore       & unlabeled          & ignore       \\
            outlier              & ignore       & 1 person           & person       \\
            car                  & car          & 2+ person          & person       \\
            bicycle              & bike         & rider              & bicyclist    \\
            bus                  & car          & car                & car          \\
            motorcycle           & car          & trunk              & trunk        \\
            on-rails             & car          & plants             & vegetation   \\
            truck                & car          & traffic sign 1     & traffic-sign \\
            other-vehicle        & car          & traffic sign 2     & traffic-sign \\
            person               & person       & traffic sign 3     & traffic-sign \\
            bicyclist            & bicyclist    & pole               & pole         \\
            motorcyclist         & bicyclist    & trashcan           & object       \\
            road                 & ground       & building           & building     \\
            parking              & ground       & cone/stone         & object       \\
            sidewalk             & ground       & fence              & fence        \\
            other-ground         & ground       & bike               & bike         \\
            building             & building     & ground             & ground       \\
            fence                & fence        &                    &              \\
            other-structure      & ignore       &                    &              \\
            lane-marking         & ignore       &                    &              \\
            vegetation           & vegetation   &                    &              \\
            trunk                & trunk        &                    &              \\
            terrain              & ground       &                    &              \\
            pole                 & pole         &                    &              \\
            traffic-sign         & traffic-sign &                    &              \\
            other-object         & object       &                    &              \\
            moving-car           & car          &                    &              \\
            moving-bicyclist     & bicyclist    &                    &              \\
            moving-person        & person       &                    &              \\
            moving-motorcyclist  & bicyclist    &                    &              \\
            moving-on-rails      & car          &                    &              \\
            moving-bus           & car          &                    &              \\
            moving-truck         & car          &                    &              \\
            moving-other-vehicle & car          &                    &             	\\ \bottomrule
        \end{tabular}}
        \caption[Class mapping for the SemanticKITTI and SemanticPOSS dataset]{Class mapping for the SemanticKITTI and SemanticPOSS dataset.}
        \label{tab:MappingKITTI-POSS}
    \end{center}
\end{table*}

\end{document}